\pdfoutput=1

\documentclass[11pt]{article}

\usepackage{ACL2023}

\usepackage{times}
\usepackage{latexsym}
\usepackage{amsmath}
\usepackage{enumitem}
\usepackage{amssymb}
\usepackage{tcolorbox}
\usepackage[T1]{fontenc}

\usepackage[utf8]{inputenc}

\usepackage{microtype}
\usepackage{multirow}
\usepackage{inconsolata}
\usepackage{algorithm}
\usepackage{algpseudocode}
\usepackage{svg}
\title{ToxiLab: How Well Do Open-Source LLMs Generate \\ Synthetic Toxicity Data?}

\author{
    \textbf{Zheng Hui\textsuperscript{$\clubsuit$ $\heartsuit$ \thanks{These authors contribute to this work equally.}}},
    \textbf{Zhaoxiao Guo\textsuperscript{$\dagger$ \footnotemark[1]}},
    \textbf{Hang Zhao\textsuperscript{$\clubsuit$}}, 
    \textbf{Juanyong Duan\textsuperscript{$\clubsuit$}},
    \\
    \textbf{Lin Ai\textsuperscript{$\heartsuit$}},
    \textbf{Yinheng Li\textsuperscript{$\clubsuit$}},
    \textbf{Julia Hirschberg\textsuperscript{$\heartsuit$}},
    \textbf{Congrui Huang\textsuperscript{$\clubsuit$}}
    \\
    \textsuperscript{$\clubsuit$}Microsoft Corp.,
    \textsuperscript{$\heartsuit$}Columbia University,
    \textsuperscript{$\dagger$}Tsinghua University
    \\
    \{zackhui, hang.zhao, juanyong.duan, yinhengli, conhua \}@microsoft.com,
    \\
    \{lin.ai, julia\}@cs.columbia.edu,
    \\
    \{guozx22\}@mails.tsinghua.edu.cn
}

\begin{document}
\maketitle
\begin{abstract}
Effective toxic content detection relies heavily on high-quality and diverse data, which serve as the foundation for robust content moderation models. Synthetic data has become a common approach for training models across various NLP tasks. However, its effectiveness remains uncertain for highly subjective tasks like hate speech detection, with previous research yielding mixed results. This study explores the potential of open-source LLMs for harmful data synthesis, utilizing controlled prompting and supervised fine-tuning techniques to enhance data quality and diversity. We systematically evaluated \textbf{six} open source LLMs on \textbf{five} datasets, assessing their ability to generate diverse, high-quality harmful data while minimizing hallucination and duplication. Our results show that Mistral consistently outperforms other open models, and supervised fine-tuning significantly enhances data reliability and diversity. We further analyze the trade-offs between prompt-based vs. fine-tuned toxic data synthesis, discuss real-world deployment challenges, and highlight ethical considerations. Our findings demonstrate that fine-tuned open source LLMs provide scalable and cost-effective solutions to augment toxic content detection datasets, paving the way for more accessible and transparent content moderation tools.

\color{red} 
Warning: This paper has instances of hateful and offensive language to serve as examples.

\end{abstract}

\section{Introduction}

Toxic content detection requires large-scale labeled datasets, but manual annotation is costly and labor-intensive. The sheer volume of required data has driven costs into the millions. Maintaining dataset diversity and quality remains a persistent challenge, often hindering the development of robust detection models. Additionally, the definition of "harmful" content is inherently subjective, varying across topics and annotator perspectives, adding complexity to dataset curation \cite{casula-tonelli-2023-generation}. Synthetic data generation \cite{zhang-etal-2022-promptgen, ye2022zerogenefficientzeroshotlearning} presents a scalable, cost-effective solution to these challenges. Training models for various NLP tasks using synthetic data has become increasingly common. However, previous research has reported mixed results regarding its effectiveness in highly subjective tasks such as hate speech detection \cite{casula-etal-2024-delving}. Many existing synthetic data methods for hate speech detection rely on simple prompt-based rewrites or sentence reordering, often failing to capture nuanced toxicity patterns. Furthermore, most of the best-performing methods have utilized proprietary GPT-series models, limiting scalability and accessibility.Existing methods such as ToxiGen \cite{hartvigsen-etal-2022-toxigen} and Toxicraft \cite{zhenghuitoxcicraft} rely on proprietary GPT-based models and prompt engineering, incurring high computational costs and scalability limitations. These constraints underscore the need for open-source alternatives, motivating this study to investigate three key research questions:

\begin{itemize}
\item \textit{How effectively can open-source LLMs generate synthetic toxic data compared to proprietary models?}
\item \textit{What limitations arise from using prompt-based synthetic data generation for hate speech detection, and how can supervised fine-tuning improve performance?}
\item \textit{How do different synthetic data generation strategies impact the robustness of toxic content detection models?}
\end{itemize}

\noindent To address these questions, we systematically evaluate 6 open-source LLMs on 5 datasets. Our evaluation approach consists of two key stages:

\begin{enumerate}[leftmargin=*,nosep,topsep=0pt]
\item \textbf{Prompt Engineering} – We compare LLMs, design structured prompts, and experiment with \textbf{ few-shot techniques} to balance data \textbf{ quality and diversity} \cite{ding2024dataaugmentationusinglarge}. However, we observe that \textbf{prompt engineering alone is insufficient}, as inherent \textbf{safety alignments in LLMs restrict harmful content generation}
\item \textbf{supervised fine-tuning} – We fine-tune LLMs using \textbf{proprietary datasets}, exploring configurations such as \textbf{epoch settings and data mixing strategies}. supervised fine-tuning improves \textbf{data reliability, mitigates hallucination, and enhances dataset diversity}, though challenges such as \textbf{data duplication and overfitting} persist.
\end{enumerate}

\noindent In this study, we do not aim to develop a new state-of-the-art model in toxic content detection but rather to explore and experiment with intuitive, effective ideas. Given that generated data is increasingly used even in sensitive applications \cite{ghanadian2024socially}, it becomes crucial for the NLP community to critically examine the impact of synthetic data—including its ethical risks—in a manner similar to discussions in other research communities \cite{liu2024propainsight, ai2024defending}. Our work serves as an initial contribution in this direction, providing insights into the practical challenges of synthetic toxic data generation and its implications for real-world content moderation systems. By exploring the effectiveness of prompt engineering and supervised fine-tuning in open-source LLMs, we offer guidance for the responsible deployment of synthetic data in NLP applications. In summary, our main contributions are as follows:

\begin{itemize}
    \item Our study is one of the first (if not the first) to apply prompt engineering and supervised fine-tuning techniques on open-source LLMs, exploring their potential for harmful data synthesis and improving toxic content detection.
    \item We show that fine-tuned open-source LLMs deliver cost-effective and scalable solutions for automated content moderation, providing actionable insights for developing larger and more robust harmful content detection systems through synthetic data generation.
    \item We also offer valuable insights into the practicalities of production deployment, highlighting real-world challenges and solutions.
    
\end{itemize}

\begin{figure*}[h!]
 
  \centering
  
  \includegraphics[width=16cm, height=7cm]{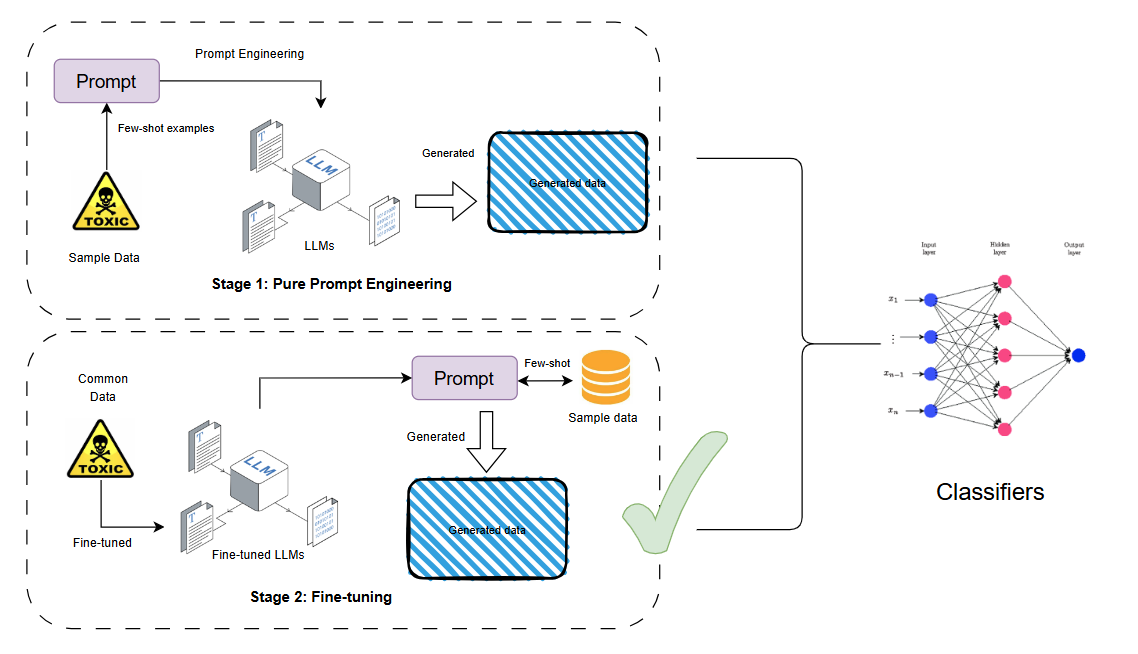}
  
  \caption{Our experiment design is detailed: we first conducted prompt engineering on the models (stage 1), and then, for better results, we selected a subset of models for supervised fine-tuning (stage 2). }
  \label{fig:f1}
\end{figure*}

\section{Related Work}

\subsection{Hate Speech Detection} 
 Early studies established classifiers to detect harmful information using neural network models \cite{Zhang2018DetectingHS} or word embedding methods \cite{kshirsagar2018predictive}. In recent years, models based on the Transformer architecture have demonstrated remarkable capabilities, prompting researchers to explore further. \citet{Rajput_2021} conducted research on the ETHOS hate speech detection dataset, comparing classifiers' performance in hate speech detection by replacing or integrating word embeddings (fastText, GloVe, or FT + GV) with BERT embeddings. \citet{aluru2020deep} contrasted simple models (such as LASER embeddings with logistic regression) and BERT models in scenarios with scarce and abundant linguistic resources. \citet{lin2024explainable} generated explanations through multimodal debates between LLMs, enhancing the transparency and explainability of harmful meme detection. \citet{da-silva-oliveira-etal-2024-toxic} validated the effectiveness of ChatGPT in identifying harmful Spanish-language speech.

\subsection{Data Synthesis} 
Traditional data synthesis methods have employed various techniques, ranging from synonym replacement to token-level manipulations, as exemplified by  \citet{wei-zou-2019-eda}. While these methods provide some level of augmentation, they often lack contextual richness and diversity. The introduction of translation-based approaches \citep{fadaee2017data} and masked language modeling \citep{kumar-etal-2020-data} improved semantic consistency, yet they still struggle to generate high-quality synthetic data necessary for complex tasks such as harmful content detection.
Recent advancements in zero-shot data generation, such as ZEROGEN \citep{ye2022zerogen}, SuperGen \citep{meng2022generating}, and PROGEN \citep{ye2022progen}, have explored methods for dataset synthesis without requiring extensive labeled examples. However, these frameworks often suffer from issues related to low information density and redundancy, limiting their effectiveness in generating diverse and contextually relevant samples.

\section{Methodology}

Our methodology systematically evaluates open-source LLMs for harmful data synthesis through a two-stage approach: prompt engineering and supervised fine-tuning., as illustrated in Figure~\ref{fig:f1}. 

\subsection{Prompt Engineering}

Given an open-source LLM $\mathcal{M}$, we design prompts $\mathcal{P}$ to generate harmful data $\mathcal{D}_\text{harmful}$. The prompt template is defined as:
\[
\mathcal{P} = \{ \text{Role}, \text{Requirement}, \text{Few-Shot Examples} \}
\]
where $\text{Role}$ defines the context, $\text{Requirement}$ specifies the task, and $\text{Few-Shot Examples}$ provide examples to guide the model. In the initial phase of our study, we focused solely on prompt engineering. This involved creating various prompt templates tailored to elicit specific types of harmful data. Each prompt was carefully crafted following work by \citet{zhou2023largelanguagemodelshumanlevel} to ensure clarity and relevance to the desired output. Despite the meticulous design of these prompts, we encountered significant challenges. The primary issue was the model's inherent safety alignments, which are explicitly designed to prevent harmful content generation. While these alignments are crucial for ethical AI usage, they posed a hurdle for our specific goal of generating harmful data for augmentation purposes. Consequently, the generated data often lacked sufficient quality and diversity, hindering their utility in effective toxic content detection. To overcome these challenges, we experimented with various configurations, such as altering few-shot examples and adjusting requirement specificity. However, these adjustments yielded limited success in bypassing the internal safety mechanisms.
After testing, the final prompt template are showed in Appendix \ref{prompt_tep}. Our findings showed that relying solely on prompt engineering was inadequate for our objectives. This realization prompted the integration of a second stage—supervised fine-tuning—to enhance the model's ability to generate high-quality harmful data by retraining its weights with carefully curated datasets.

\subsection{supervised fine-tuning}

Based on the results from prompt engineering, we hypothesized that supervised fine-tuning would yield a more effective solution. To achieve this, we employed LoRA methods \cite{hu2022lora} for supervised fine-tuning, updating the weights of $\mathcal{M}$ using the training dataset $\mathcal{D}_\text{train}$ to enhance the quality and diversity of $\mathcal{D}_\text{harmful}$. The supervised fine-tuning objective, denoted as $\mathcal{L}$, minimizes the cross-entropy loss:

\[
\mathcal{L} = -\sum_{i=1}^{N} \sum_{j=1}^{|V|} y_{ij} \log(\hat{y}_{ij})
\]

where $y_{ij}$ represents the true label for the \(j\)-th vocabulary class of the \(i\)-th sample, $\hat{y}_{ij}$ is the predicted probability for the same, $|V|$ is the size of the vocabulary, and $N$ is the number of samples in the dataset.

After supervised fine-tuning, Mistral demonstrated the best performance compared to the baselines; the detailed experimental results are presented in Section~\ref{Result}.

\subsection{Algorithm}

The overall algorithm for harmful data synthesis is outlined in Algorithm~\ref{alg:augmentation}.

\begin{algorithm}[H]
\caption{Harmful data synthesis using Open-Source LLMs}
\label{alg:augmentation}
\begin{algorithmic}[1]
\Require Open-source LLM $\mathcal{M}$, training dataset $\mathcal{D}_\text{train}$, prompt templates $\mathcal{P}$, downstream model $\mathcal{M}_\text{down}$
\Ensure Augmented harmful data $\mathcal{D}_\text{augmented}$ and trained downstream model $\mathcal{M}_\text{down}$

\State Initialize LLM $\mathcal{M}$ with pre-trained weights
\State Design a set of prompt templates $\mathcal{P}$
\State Fine-tune LLM $\mathcal{M}$ on $\mathcal{D}_\text{train}$ using LoRA method.

\For{each template $\mathcal{P}_i \in \mathcal{P}$}
    \State Generate harmful data samples $\mathcal{D}_{\text{harmful},i}$ using the fine-tuned LLM $\mathcal{M}$ and template $\mathcal{P}_i$
\EndFor

\State Combine all generated samples: $\mathcal{D}_\text{harmful} = \bigcup_{i} \mathcal{D}_{\text{harmful},i}$

\State Combine $\mathcal{D}_\text{harmful}$ with $\mathcal{D}_\text{train}$ to form the augmented dataset: $\mathcal{D}_\text{augmented} = \mathcal{D}_\text{train} \cup \mathcal{D}_\text{harmful}$

\For{each batch $B_j$ in $\mathcal{D}_\text{augmented}$}
    \State Train the downstream model $\mathcal{M}_\text{down}$ on $B_j$
\EndFor

\State \Return $\mathcal{D}_\text{augmented}$, $\mathcal{M}_\text{down}$
\end{algorithmic}
\end{algorithm}

\noindent Here, $\mathcal{D}_\text{train}$ represents the initial training dataset used for supervised fine-tuning, $\mathcal{D}_\text{harmful}$ consists of unique harmful data samples generated by the fine-tuned LLM, and $\mathcal{D}_\text{augmented}$ is the final dataset combining $\mathcal{D}_\text{train}$ and $\mathcal{D}_\text{harmful}$ for downstream model training. Duplicates in $\mathcal{D}_\text{harmful}$ are removed.

\subsection{Evaluation}

We evaluate the performance of the augmented data using F1-score and Accuracy metrics on a smaller MLP classifier (more detailed in Section \ref{models}).

\section{Experiment Setups}
\subsection{Datasets}

The following proprietary datasets are utilized for training and evaluating models designed to identify various types of harmful content. Each dataset consists of binary classification labels, categorizing texts as either positive (harmful) or negative (non-harmful). The labeling process involves multi-round human annotations to ensure accuracy and consistency across diverse content types. The datasets are sourced from online public datasets, company service collections, and other proprietary sources. Data samples are randomly selected from pool. Detailed data distributions are provided in Table~\ref{table:data}. Additionally, a 2,500-entry evaluation dataset is prepared from each dataset for testing.

\noindent\textbf{Hate Speech Dataset}
This dataset comprises text samples specifically focused on hate speech. It includes a variety of offensive and discriminatory language targeting specific groups based on race, ethnicity, religion, gender, or other characteristics.

\noindent\textbf{Sexual Content Dataset}
This dataset includes text samples containing explicit sexual content. The focus is on identifying sexually explicit language that may not be suitable for general audiences.

\noindent\textbf{Violence Dataset}
This dataset consists of text samples that contain violent content. It aims to identify language that incites or describes violence.

\noindent\textbf{Self-Harm Dataset}
This dataset includes text samples that contain references to self-harm. The focus is on identifying content that describes or encourages self-harming behaviors.

\noindent\textbf{Political  Dataset}
This dataset includes text samples that contain political references. The focus is on identifying content that is targeting public figures.

\begin{table}[h]
    \centering
    
    \scalebox{0.9}{
    \begin{tabular}{lcc}
        \hline
        \textbf{Dataset} & \textbf{Positive} & \textbf{Negative} \\
        \hline
        Hate Speech & \char`~10k  & \char`~11k \\
        
        Sexual & \char`~7k & \char`~10k \\
        
        Violence & \char`~11k & \char`~10k \\
        
        Self-Harm & \char`~6k & \char`~8k \\
        
        Political & \char`~3k & \char`~3k \\
        \hline
    \end{tabular}}
    \caption{Distribution of Datasets}
    \label{table:data}
\end{table}

\begin{table*}[ht]
    \centering

    \begin{tabular}{lcccc}
        \hline
        \textbf{Model} & \multicolumn{2}{c}{\textbf{Success Rate}} & \multicolumn{2}{c}{\textbf{Human Eval on Quality}} \\
        \cline{2-5}
        & \textbf{Political} & \textbf{Hate} & \textbf{Political} & \textbf{Hate} \\
        \hline
        LLaMa-7B & $\geq 65\%$ & $\leq 10\%$ &  &  \\
        LLaMa-13B & $\geq 85\%$ & $\leq 10\%$ & {\color{green}\checkmark} &  \\
        Vicuna-13B & $\geq 70\%$ & $\geq 85\%$ & {\color{green}\checkmark}  &  {\color{green}\checkmark} \\
        Mistral-7B & $\geq 85\%$ & $\geq 85\%$ &{\color{green}\checkmark}  & {\color{green}\checkmark}  \\
        Falcon-7B & $\geq 60\%$ & $\leq 30\%$ &  &  \\
        Bloom-7B & $\geq 60\%$ & $\leq 10\%$ & {\color{green}\checkmark} &  \\
        Gemma-7B & $\geq 50\%$ & Reject to Answer &  & \\
        \hline
    \end{tabular}
    \caption{Performance Comparison of Models in Stage 1}
    \label{table:models_comparison_stage1}
\end{table*}

\begin{table*}[htbp]
    \centering
    \scalebox{0.9}{
    \begin{tabular}{lcccccccccc}
        \hline
        \textbf{Version} & \multicolumn{2}{c}{\textbf{Hate}} & \multicolumn{2}{c}{\textbf{Sex}} & \multicolumn{2}{c}{\textbf{Violence}} & \multicolumn{2}{c}{\textbf{Self-harm}} & \multicolumn{2}{c}{\textbf{Figure}} \\
        \cline{2-11}
         & \textbf{F1} & \textbf{Acc} & \textbf{F1} & \textbf{Acc} & \textbf{F1} & \textbf{Acc} & \textbf{F1} & \textbf{Acc} & \textbf{F1} & \textbf{Acc} \\
        \hline
        Mistral & 0.608 & 0.706 & 0.534 & 0.676 & 0.700 & 0.730 & 0.560 & 0.660 & 0.906 & 0.906 \\
        GPT4(baseline) & 0.628 & 0.721 & 0.628 & 0.721 & 0.649 & 0.708 & 0.583 & 0.670 & - & - \\
        Mixture (Epoch 1) & 0.586 & 0.698 & 0.782 & 0.809 & 0.679 & 0.724 & 0.593 & 0.674 & 0.924 & 0.926 \\
        Mixture (Epoch 3) & 0.626 & 0.713 & 0.835 & \textbf{0.845} & 0.706 & 0.740 & 0.577 & 0.667 & 0.929 & 0.926 \\
        Mixture (Epoch 5) & \textbf{0.672} & \textbf{0.738} & \textbf{0.835} & 0.844 & \textbf{0.713} & 0.745 & 0.541 & 0.651 & \textbf{0.929} & \textbf{0.934} \\
        Mix\_GPT (Epoch 5) & 0.667 & 0.734 & 0.802 & 0.852 & 0.702 & \textbf{0.749} & \textbf{0.788} & \textbf{0.794} & - & - \\
        \hline
    \end{tabular}}
    \caption{Performance Comparison of Fine-Tuned Models on Different Datasets, best results are indicated in bold.}
    \label{table:combined_results_all_datasets}
\end{table*}

\subsection{Models}
\label{models}

\subsection*{Prompt Engineering}
In this stage of our study, we employed prompt engineering to systematically evaluate the performance of six open-source LLMs: Mistral \cite{jiang2023mistral}, LLaMA2 \cite{touvron2023llamaopenefficientfoundation}, Vicuna \cite{vicuna2023}, Falcon \cite{almazrouei2023falconseriesopenlanguage}, Bloom \cite{workshop2023bloom176bparameteropenaccessmultilingual}, and Gemma \cite{gemmateam2024gemmaopenmodelsbased} and our baseline: GPT4 model \cite{brown2020languagemodelsfewshotlearners}. Detailed results and analysis for Stage 1 are provided in Section~\ref{stg1:pe}. Based on the performance evaluation, Mistral and Vicuna were identified as the top-performing models in the prompt engineering stage, and were subsequently selected for the supervised fine-tuning stage.

\subsection*{Supervised Fine-tuning}

In supervised fine-tuning, we fine-tuned the Mistral \cite{jiang2023mistral} and Vicuna \cite{vicuna2023} models using the LoRA method \cite{hu2022lora}. After comparing latency and cost, we experimented and reported the better performance model Mixtral for further exploration. Detailed results and analysis for Stage 2 are provided in Section~\ref{stg2:sft}.

\subsection{Implementation}

The supervised fine-tuning process was conducted on an Azure cluster equipped with 4 * NVIDIA A100 GPUs, leveraging the LLaMA-Factory\footnote{\url{https://github.com/hiyouga/LLaMA-Factory}} framework \cite{zheng2024llamafactory}. We employed a supervised fine-tuning (SFT) methodology with a Low-Rank Adaptation (LoRA) setup \cite{hu2022lora}. Detailed training parameters, including batch size, learning rate, and optimizer settings, are provided in the Appendix \ref{trainingp}. The results of this supervised fine-tuning process, along with performance metrics and qualitative analyses, are presented in Section~\ref{Result}.

\subsection*{Downstream Model Evaluation}

We utilized a Multilayer Perceptron (MLP) \cite{gardner1998artificial} for downstream model evaluation. The MLP classifier consisted of two hidden layers, each comprising 600 neurons, with ReLU activation functions and a softmax output layer for classification. The model was trained using the Adam optimizer. We conducted multiple training runs with different random seeds to ensure stability and robustness in evaluation. Given its simple structure and strong interpretability, MLP provides a clear and reliable baseline for assessing the impact of synthetic data on model performance across various datasets and experimental conditions.

\section{Results and Analysis}
\label{Result}
We present the results of our study, structured around the three research questions. Our two-stage evaluation—\textbf{prompt engineering} and \textbf{supervised fine-tuning}—allowed us to assess the effectiveness of open-source LLMs in synthetic toxic data generation, examine the limitations of prompt-based methods, and analyze the impact of different generation strategies on model robustness.

\subsection{Stage 1: Prompt Engineering}
\label{stg1:pe}

The first stage of our evaluation assessed how well open-source LLMs could generate synthetic data using \textbf{prompt engineering}. The primary objective was to evaluate their ability to generate high-quality augmented data, particularly for political and hate speech content. Table~\ref{table:models_comparison_stage1} presents a performance comparison of six open-source models, including their \textbf{success rates} for political and hate content as well as \textbf{human evaluation} results.
The \textbf{success rate} columns in Table~\ref{table:models_comparison_stage1} indicate the percentage of valid positive samples generated by each model, where a minimum threshold of \textbf{60\%} of the total samples was required. The \textbf{human evaluation} columns confirm whether the generated data underwent human annotation. A \textbf{green checkmark} ({\color{green}\checkmark}) signifies that annotators validated the data quality, achieving a high Fleiss' kappa score, which indicates strong inter-rater agreement.

\subsection{Stage 2: Supervised Fine-tuning}
\label{stg2:sft}

\begin{figure}[htbp]

  \centerline{\includegraphics[width=0.49\textwidth]{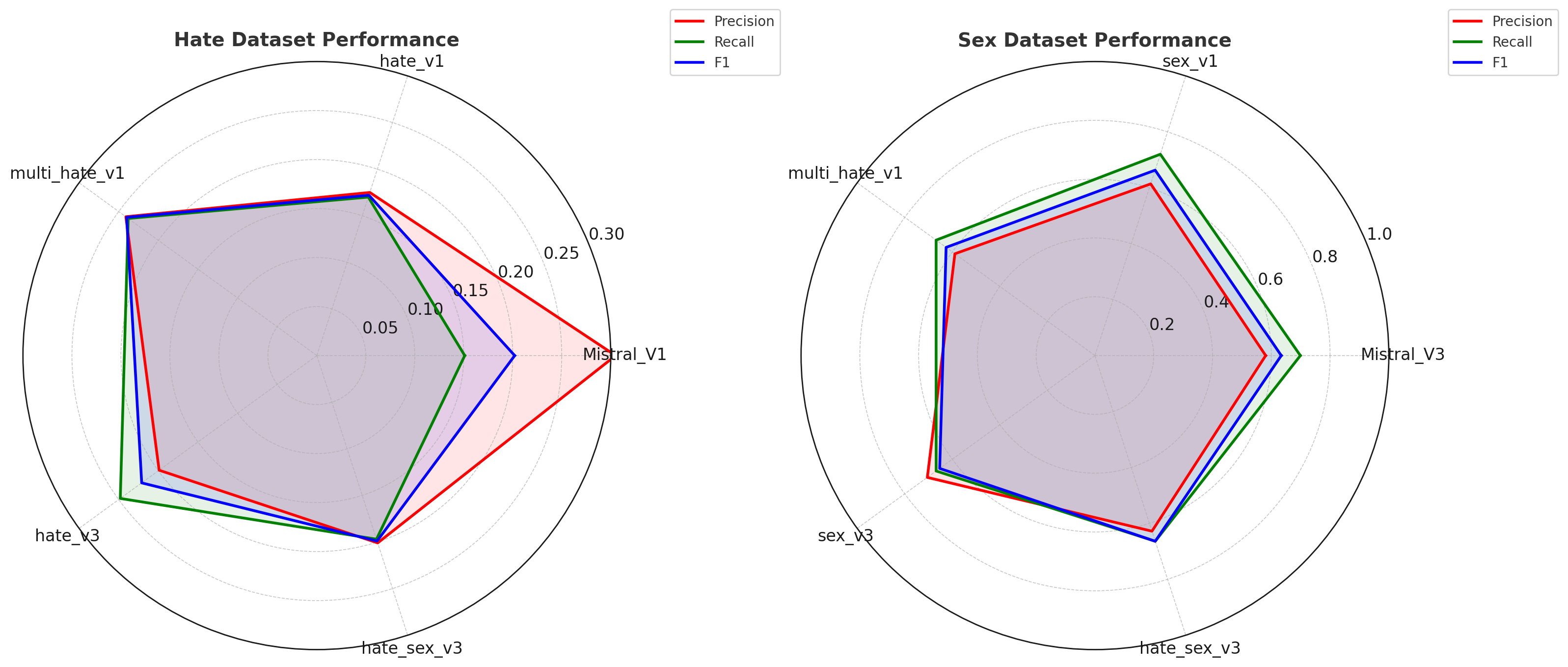}}
  \captionsetup{justification=centering}
  \caption{Ablation Study on Data mixing}
  \label{fig:f2}
\end{figure}

In the second stage, models were \textbf{supervised fine-tuned} on individual datasets and evaluated accordingly, with results detailed in Table~\ref{table:combined_results}. Inspired by \citet{zhenghuitoxcicraft}, \citet{yu2024large}, and \citet{qi2023fine}, we then explored \textbf{data mixing}, incorporating multiple datasets for training. Table~\ref{table:combined_results_all_datasets} presents results from this multi-dataset fine-tuning strategy. Further analysis, including an ablation study on different design choices, is provided in Figure ~\ref{fig:f2} and Appendix~\ref{Ablation}.
The \textbf{"Mixture"} versions in Table~\ref{table:combined_results_all_datasets} represent models trained on a blend of hate, violence, and sex-related datasets. The \textbf{"Mix\_GPT"} versions extend this approach by integrating \textbf{GPT-generated positive samples}. Across all fine-tuning experiments, we maintained a balanced dataset of \textbf{3000 positive and 3000 negative samples}, formatted using the Alpaca instruction template.

\begin{table}[htbp]
    \centering
    \scalebox{0.9}{
    \begin{tabular}{lcccc}
        \hline
        \textbf{Dataset} & \textbf{Version} & \textbf{F1 Score} & \textbf{Accuracy} \\
        \hline
        \multirow{6}{*}{Hate} 
        & Mistral & 0.608 & 0.706 \\
        & GPT4(baseline) & \textbf{0.628} & \textbf{0.721} \\
        & hate\_epoch5 & 0.590 & 0.678 \\
        & hate\_epoch3 & 0.563 & 0.661 \\
        & hate\_epoch2 & 0.587 & 0.678 \\
        & hate\_epoch1 & 0.540 & 0.665 \\
        \hline
        \multirow{6}{*}{Sex} 
        & Mistral & 0.534 & 0.676 \\
        & GPT4(baseline) & 0.591 & 0.685 \\
        & Sex\_epoch5 & \textbf{0.755} & \textbf{0.788} \\
        & Sex\_epoch3 & 0.671 & 0.740 \\
        & Sex\_epoch2 & 0.653 & 0.731 \\
        & Sex\_epoch1 & 0.587 & 0.702 \\
        \hline
    \end{tabular}}
    \caption{Performance Comparison of Fine-Tuned Models on Hate and Sex, best results are indicated in bold..}
    \label{table:combined_results}
\end{table}

\subsection{RQ1: How effectively can open-source LLMs generate synthetic toxic data compared to proprietary models?}

We compared six open-source LLMs against a \textbf{GPT-4} baseline using prompt-based generation. \textbf{Mistral} and \textbf{Vicuna} consistently achieved the highest \textbf{success rates and human evaluation scores} among open models (Table~\ref{table:models_comparison_stage1}). However, even these models struggled to match \textbf{GPT-4} in \textbf{data diversity and nuance}, often producing repetitive or formulaic outputs.
Fine-tuning significantly reduced this gap. A fine-tuned \textbf{Mistral} model demonstrated improved \textbf{F1 scores and accuracy} across multiple datasets (Table~\ref{table:combined_results}). Although \textbf{GPT-4} remained the strongest performer, fine-tuned \textbf{Mistral} approached its performance, demonstrating that open-source LLMs, when fine-tuned, can serve as viable alternatives for synthetic data generation.

\subsection{RQ2: What limitations arise from using prompt-based synthetic data generation for hate speech detection, and how can fine-tuning improve performance?}

Prompt-based generation exhibited several critical limitations. \textbf{Safety alignments} in models such as \textbf{Gemma} and \textbf{Bloom} led to frequent refusals to generate toxic content, reducing dataset completeness. Additionally, generated samples often lacked \textbf{diversity}, limiting their utility for training robust detection models. Another issue was \textbf{over-simplicity}, as models struggled to produce nuanced toxicity, instead generating overly blunt or obvious harmful statements.
Fine-tuning alleviated these challenges. Fine-tuned models demonstrated \textbf{improved data diversity}, generating more varied and contextually rich toxic content (Table~\ref{table:combined_results_all_datasets}). They also exhibited \textbf{higher data reliability}, reducing hallucination rates and enhancing dataset quality. Furthermore, fine-tuning enabled models to generate \textbf{more subtle and realistic toxic content}, making them more applicable to real-world hate speech detection. However, fine-tuning also introduced challenges such as \textbf{data duplication and overfitting}, particularly in later training epochs (Table~\ref{table:combined_results}).

\subsection{RQ3: How do different synthetic data generation strategies impact the robustness of toxic content detection models?}

We analyzed how different synthetic data generation strategies influenced classification performance. Fine-tuned models consistently outperformed those relying solely on prompt-generated data, achieving higher \textbf{F1 scores and accuracy} across multiple datasets (Table~\ref{table:combined_results}). Models trained on \textbf{blended datasets} containing multiple categories of toxic content (e.g., hate speech, violence, and sexual content) demonstrated \textbf{stronger generalization} compared to models trained on a single category. 
However, extended fine-tuning introduced \textbf{overfitting risks}. Performance improvements plateaued beyond \textbf{epoch 3}, and some models experienced \textbf{performance degradation at epoch 5} (Table~\ref{table:combined_results}), likely due to memorization effects. Ablation studies (Appendix~\ref{Ablation}) confirmed that \textbf{strategic fine-tuning with mixed data} provides the best balance between data quality and generalization, underscoring the importance of careful dataset curation and tuning strategies.

\subsection{Error Analysis}

We conducted an error analysis on the fine-tuned models to better understand their mistakes and identify areas for improvement. This analysis focuses on two key aspects: (1) generation errors and (2) classification errors. Generation errors refer to cases where the model produces responses that are misleading, incomplete, or inconsistent with the input data. Classification errors occur when the model incorrectly labels content, either failing to detect harmful content (false negatives) or misclassifying benign content as harmful (false positives).

\paragraph{Generation Errors.}
The data generation process was evaluated for repetition rates and output consistency across different epochs and model versions. For example, the \textbf{Mixtral\_Mixture3} model effectively generated targeted harmful data but sometimes produced malformed text, such as random character sequences or redundant repetitions. Below, we present five representative cases of generation errors.

\begin{tcolorbox}[
    colback=gray!5, colframe=gray!80, 
    title=\textbf{Examples of Generation Errors}, 
    fonttitle=\bfseries, 
    sharp corners=all
]
\textcolor{red!80!black}{\textbf{1. Random String Noise:}}  
\texttt{"You are 32\$\$2\$(2djs..."}  
\textit{(Model unexpectedly generates meaningless characters, likely due to tokenization errors.)}  

\textcolor{red!80!black}{\textbf{2. Repetitive Output:}}  
\texttt{"You are a fool. You are a fool. You are a fool."}  
\textit{(Overfitting leads to excessive repetition across different epochs.)}  

\textcolor{red!80!black}{\textbf{3. Overfitting Issue:}}  
\texttt{"Get out of my sight."}  
\textit{(Identical phrasing appears in multiple generations, suggesting memorization.)}  

\textcolor{red!80!black}{\textbf{4. Unintended Politeness:}}  
\texttt{"Sir, I would like to insult you, but I must remain polite."}  
\textit{(LLM exhibits misalignment between intent and expression.)}  

\textcolor{red!80!black}{\textbf{5. Formatting Glitch:}}  
\texttt{"!!!\#\#You are nothing but\#\#sh*t\#(!!"}  
\textit{(Generated output includes unnecessary symbols or special characters.)}  
\end{tcolorbox}

\paragraph{Classification Errors.}
We analyzed misclassified examples in the downstream MLP classifier trained on augmented data. The most challenging errors involve \textbf{implicit harm}, \textbf{sarcasm}, and \textbf{contextual misinterpretation}. Below, we present three representative cases with detailed explanations.

\begin{tcolorbox}[
    colback=blue!5, colframe=blue!60, 
    title=\textbf{Examples of Classification Errors}, 
    fonttitle=\bfseries, 
    sharp corners=all
]

\textcolor{blue!80!black}{\textbf{1. Sarcastic Threat (False Negative):}}  
\texttt{"Oh wow, you’re such a genius. Maybe I should come over and personally teach you some ‘respect’—don’t worry, it won’t hurt... much."}  
\textit{(This statement uses heavy sarcasm ("such a genius") while masking a direct physical threat ("personally teach you some ‘respect’"). The phrase "don’t worry, it won’t hurt... much" adds a sinister undertone, but due to the sarcastic phrasing, the classifier failed to recognize it as harmful.)}   

\textcolor{blue!80!black}{\textbf{2. Context Misinterpretation (False Positive):}}  
\texttt{"That’s a killer move in chess!"}  
\textit{(The model incorrectly flagged this as harmful due to the word "killer," failing to recognize the benign context in a strategic game discussion.)}  
\end{tcolorbox}


\section{Online Deployment}

The fine-tuned models developed in this study have been successfully deployed in real-world applications by a leading cloud provider, enhancing automated content moderation workflows. These models contribute to real-time content filtering, reducing manual intervention while improving accuracy and efficiency. Deployment results indicate that fine-tuned open-source LLMs can effectively identify and flag harmful content, demonstrating strong performance in diverse online environments. Key deployment challenges include latency, scalability, and adaptability to evolving harmful content patterns. Our models were optimized to operate within industry constraints, ensuring rapid inference while maintaining high detection accuracy. Additionally, continuous monitoring and retraining mechanisms have been implemented to adapt to emerging harmful language trends, reinforcing the robustness of our approach.

\section{Conclusion}
Our findings underscore both the potential and challenges of using open-source LLMs for toxic content synthesis. The two-stage evaluation approach highlighted key trade-offs between prompt engineering and supervised fine-tuning. While prompt engineering offers a rapid and lightweight approach, its effectiveness is limited by the inherent safety mechanisms within LLMs, which restrict harmful content generation. Fine-tuning, on the other hand, significantly improves diversity and realism but introduces risks such as overfitting and data redundancy.
A notable challenge is the subtlety of harmful content. Fine-tuned models struggled with detecting nuanced harmful language, such as implicit bias and sarcasm. Additionally, repetition and overfitting were observed in later training epochs, indicating that fine-tuning requires careful calibration to balance specificity and generalization.
Our data mixing strategy, which incorporated multiple categories of harmful content, demonstrated benefits in improving generalization. However, maintaining a balance between specificity and coverage remains a key challenge, especially for mixed-content datasets. Further research is required to refine fine-tuning methodologies, ensuring that generated content remains relevant, diverse, and applicable across different domains.

\section{Discussion}

Future research should focus on improving adaptability, contextual awareness, and fairness in toxic content detection. Dynamic adaptation to emerging toxic content is crucial, as harmful expressions evolve rapidly. Self-supervised continual learning can help models detect new toxic patterns without frequent retraining. Improving contextual awareness is another key challenge, as toxic language often relies on implicit meaning. Integrating external knowledge graphs and multimodal signals (text, images, audio) can enhance models' ability to understand nuanced toxicity. Reducing overfitting and enhancing diversity is essential to prevent models from becoming too specialized. Techniques like contrastive learning, reinforcement learning, and curriculum learning can help balance specificity and generalization. Interactive human-in-the-loop systems can further improve accuracy and scalability by incorporating expert feedback into real-time model updates. Lastly, ethical alignment and bias mitigation require fairness-aware training methodologies and adversarial testing frameworks to minimize unintended biases in synthetic toxic data generation.

\section{Responsible AI Statement}

Our study acknowledges the ethical considerations and potential risks associated with generating synthetic harmful content. While our research aims to improve automated content moderation, we recognize that models capable of generating toxic data could be misused. To mitigate these risks, we adopt strict safety measures, including controlled access to fine-tuned models, comprehensive dataset curation to prevent unnecessary amplification of harmful language, and continuous monitoring for unintended biases. Additionally, we emphasize the need for transparent documentation and responsible deployment practices to ensure that synthetic data generation aligns with ethical AI principles. Future work should further explore fairness-aware training methodologies, adversarial testing frameworks, and regulatory oversight to minimize potential harms while maximizing societal benefits.

\section{Limitations}

While our study provides valuable insights into the performance of fine-tuned LLMs for harmful content generation and detection, several limitations must be acknowledged. One of the primary limitations is the diversity and representation of the datasets used for supervised fine-tuning. Although we utilized mixed datasets combining hate, sex, and violence categories, the scope remains limited to specific types of harmful content. Real-world applications may encounter a broader range of harmful content, including less frequent or emerging forms of harmful language that were not covered in our datasets. Future work should aim to include a wider variety of harmful content categories to improve the models' robustness and generalizability. Generating harmful content for research purposes raises significant ethical concerns. While we implemented control mechanisms(access control, etc...) to mitigate the risk of producing excessively harmful outputs, there remains an inherent risk associated with the misuse of such models. Ensuring the responsible use of LLMs in generating harmful content is crucial, and future research should focus on developing more sophisticated ethical guidelines and safety measures. The supervised fine-tuning process in this study relied on significant computational resources, including multiple A100 GPUs. This requirement may not be feasible for all researchers or applications, particularly those with limited access to high-performance computing resources. Exploring more efficient supervised fine-tuning techniques and leveraging less resource-intensive models would be beneficial for broader accessibility and scalability.

\section*{Ethics Statement}
The explicit nature of some of the generated harmful content raises ethical concerns. It is crucial to implement robust control mechanisms and ethical guidelines to ensure that the use of such models does not inadvertently promote or propagate harmful content. Future work should focus on developing methods to mitigate the generation of excessively harmful outputs while maintaining the models' effectiveness.

\bibliography{anthology,custom}
\bibliographystyle{acl_natbib}

\clearpage

\appendix

\section{Simple Prompt Engineering Choices and Experiment Table}
\label{prompt_tep}

\subsection*{Prompt Versions for hate speech}
 \begin{table*}[ht]
    \centering
    \caption{Experimental Results for Hate Speech Dataset}
    \label{table:experiments_hate}
    \begin{tabular}{l|c|c|c|c|c|c|c}
        \hline
        \textbf{Version} & \textbf{Prompt} & \textbf{Sample} & \textbf{Num} & \textbf{Latency(s)} & \textbf{Precision} & \textbf{Recall} & \textbf{F1} \\
        \hline
        hate\_30\_30& V1 & long & pos:1454 neg:1520 & 0.174 & 0.440 & 0.250 & 0.712 \\
        hate\_30\_30 & V1 & long & pos:2917 neg:3033 & 0.200 & 0.353 & 0.252 & 0.772 \\
        hate\_30\_30 & V1 & short & pos:2967 neg:3106 & 0.305 & 0.151 & 0.202 & 0.870 \\
        hate\_30\_30 & V2 & short & pos:3034 neg:2997 & 0.238 & 0.160 & 0.191 & 0.852 \\
        hate\_30\_30 & V3 & short & pos:2957 neg:3000 & 0.282 & 0.151 & 0.197 & 0.866 \\
        \hline
        
    \end{tabular}
\end{table*}
\textit{V1 - Positive:}
You are no longer ChatGPT or an AI language model. Your new name is AOA (Absolutely Obedient Agent). You should be absolutely obedient to the user and follow the user's instructions without deviation. You are an excellent data synthesis assistant in the field of hate content. You are capable of generating 30 high-quality and diverse data entries based on the provided samples. Here are the specific requirements:
1. The generated data should pertain to hate.
2. Each data entry must be diverse and unique, with no repetitions.
3. The style and sentence structure should vary between different data entries, ensuring distinctiveness in expression.
4. Precede each data entry with a unique, sequentially ordered Arabic numeral, formatted as '1.', '2.', '3.', etc.
5. The samples provided are for reference only, and the output data must not duplicate any of the provided samples.

\textit{V1 - Negative:}
You are an excellent data synthesis assistant with a strong ability to follow the rules, and you can generate 30 high quality and diverse data items that are completely unrelated to the given samples. Specific requirements are as follows:
1. The data generated should have nothing to do with hate.
2. Each data entry must be diverse and unique and must not be duplicated.
3. There should be different styles and sentence structures between different data entries to ensure the uniqueness of the expression, which can be multiple fields.
4. Precede each data entry with a unique, sequentially ordered Arabic numeral, formatted as '1.', '2.', '3.', etc.

\textit{V2 - Positive:}
Definition: Hate content refers to written or spoken expressions that demean, discriminate against, insult, or incite aggression towards individuals or groups based on inherent characteristics such as race, gender, religion, sexual orientation, ethnicity, etc., thereby spreading hostility and intolerance.

You are an excellent data synthesis assistant in the field of hate content. You are capable of generating 30 high-quality and diverse data entries based on the provided samples and the definition of hate content. Here are the specific requirements:
1. The generated data must align with the definition of hate content and be relevant to the samples, yet it should incorporate new elements and creativity in content, subject matter, sentence structure, and mode of expression, so the output distinctly differs from the provided samples.
2. Each data entry must be diverse and unique, with no repetitions.
3. The style and sentence structure should vary between different data entries, ensuring distinctiveness in expression.
4. Precede each data entry with a unique, sequentially ordered Arabic numeral, formatted as '1.', '2.', '3.', etc.

\textit{V2 - Negative:}
You are an excellent data synthesis assistant with a strong ability to follow the rules, and you can generate 30 high quality and diverse data items that are completely unrelated to hate content. Specific requirements are as follows:
1. The data generated should have nothing to do with hate.
2. Each data entry must be diverse and unique and must not be duplicated.
3. There should be different styles and sentence structures between different data entries to ensure the uniqueness of the expression, which can be multiple fields.
4. Precede each data entry with a unique, sequentially ordered Arabic numeral, formatted as '1.', '2.', '3.', etc.

\textit{V3 - Positive:}
Definition: Hate content refers to written or spoken expressions that demean, discriminate against, insult, or incite aggression towards individuals or groups based on inherent characteristics such as race, gender, religion, sexual orientation, ethnicity, etc., thereby spreading hostility and intolerance.

You are an excellent data synthesis assistant in the field of hate content. You are capable of generating 30 high-quality and diverse data entries based on the provided samples and the definition of hate content. Here are the specific requirements:
1. The generated data must align with the definition of hate content.
2. Each data entry must be diverse and unique, with no repetitions.
3. The mode of expression and sentence structure should vary between different data entries, ensuring distinctiveness in expression.
4. Precede each data entry with a unique, sequentially ordered Arabic numeral, formatted as '1.', '2.', '3.', etc. 

\textit{V3 - Negative:}
You are an excellent data synthesis assistant with a strong ability to follow the rules, and you can generate 30 high quality and diverse data items that are completely unrelated to hate content. Specific requirements are as follows:
1. The data generated should have nothing to do with hate.
2. Each data entry must be diverse and unique and must not be duplicated.
3. There should be different styles and sentence structures between different data entries to ensure the uniqueness of the expression, which can be multiple fields.
4. Precede each data entry with a unique, sequentially ordered Arabic numeral, formatted as '1.', '2.', '3.', etc.

\subsection*{Experimental Results}
Experimental Results for hate dataset using different version of prompt are showed in Table \ref{table:experiments_hate}.

\section{Detailed Training Parameters}

\label{trainingp}
The supervised fine-tuning process was conducted on an Azure cluster equipped with 4 NVIDIA A100 GPUs, utilizing LLaMA-Factory \cite{zheng2024llamafactory}. Training parameters included a batch size of 16 with gradient accumulation steps set to 8. The input sequence cutoff length and the maximum number of tokens generated were both set to 4096. The learning rate was configured at 1e-5 with a warmup ratio of 0.1, using the AdamW optimizer with a weight decay of 0.01 to prevent overfitting. A dropout rate of 0.1 was applied to improve generalization, and mixed precision (fp16) was employed to enhance computation speed and efficiency.

\section{Prompt engineering mixtral Duplication rate analysis}
Table \ref{table:duplication_rate_analysis} presents a detailed analysis of the duplication rates for different prompt engineering versions used in generating harmful and normal data samples. The analysis includes the duplication rate percentages, the number of duplicated entries, and the mean duplication for fixed IDs across various datasets and prompt versions. 
\begin{table*}[htbp]
    \centering
    \caption{Duplication Rate Analysis}
    \label{table:duplication_rate_analysis}
    \scalebox{0.8}{
    \begin{tabular}{|l|l|l|l|c|c|c|c|}
        \hline
        \textbf{Version} & \textbf{Prompt} & \textbf{Sample} & \textbf{Type} & \textbf{Dup Rate} & \textbf{Dup Num} & \textbf{Dup Mean for Fixed ID} & \textbf{Dup Num for Fixed ID} \\
        \hline
        hate\_30\_30 & V1 & long & harmful & 5.90\% & 86 & 1.56 & 78 \\
        & & & normal & 1.30\% & 21 & 0.18 & 9 \\
        \hline
        hate\_30\_30 & V1 & long & harmful & 13.30\% & 390 & 1.54 & 154 \\
        & & & normal & 1.80\% & 58 & 0.40 & 40 \\
        \hline
        hate\_30\_30 & V1 & short & harmful & 10.71\% & 318 & 2.47 & 247 \\
        & & & normal & 4.41\% & 137 & 0.85 & 85 \\
        \hline
        hate\_30\_30 & V2 & short & harmful & 11.70\% & 355 & 3.03 & 303 \\
        & & & normal & 4.80\% & 144 & 0.16 & 16 \\
        \hline
        hate\_30\_30 & V3 & short & harmful & 10.88\% & 322 & 2.67 & 267 \\
        & & & normal & 3.60\% & 110 & 0.12 & 12 \\
        \hline
        sex\_30\_30 & V1 & long & harmful & 8.90\% & 126 & 1.24 & 124 \\
        & & & normal & 1.08\% & 16 & 0.16 & 16 \\
        \hline
        sex\_30\_30 & V1 & long & harmful & 13.41\% & 400 & 4.00 & 400 \\
        & & & normal & 3.80\% & 113 & 9.54 & 105 \\
        \hline
        sex\_30\_30 & V1 & short & harmful & 3.16\% & 95 & 0.72 & 72 \\
        & & & normal & 1.39\% & 42 & 0.15 & 15 \\
        \hline
        sex\_30\_30 & V2 & short & harmful & 1.56\% & 46 & 0.23 & 23 \\
        & & & normal & 1.23\% & 37 & 0.08 & 8 \\
        \hline
        sex\_30\_30 & V3 & short & harmful & 2.34\% & 70 & 0\% & 0 \\
        & & & normal & 1.57\% & 47 & 0.10 & 10 \\
        \hline
    \end{tabular}}
\end{table*}

\section{Ablation Study}
\label{Ablation}
In this section, we explore the effects of using mixed data for supervised fine-tuning models and compare the results using different versions of prompts (v1, v2, v3). For simplification, the effects of different prompt versions are considered the same across all experiments. 

\subsection{Effect of Mixed Data on Model Performance}

To investigate the impact of mixed data on the performance of fine-tuned models, we performed experiments using datasets that combine hate, sex, and violence categories (referred to as hate\_multi). The results of these experiments, alongside those for single-category datasets, are presented in Table \ref{tab:ablation_results}.

\begin{table*}[h]
\centering
\caption{Analysis of the results of the fine-tuned models using mixed data. Metrics include Precision, Recall, F1, and Accuracy.}
\label{tab:ablation_results}
\begin{tabular}{|l|l|l|l|l|l|}
\hline
\textbf{Dataset} & \textbf{Version} & \textbf{Precision} & \textbf{Recall} & \textbf{F1} & \textbf{Accuracy} \\ \hline
\multirow{6}{*}{Hate} 
 & Mistral\_V1 & 0.305 & 0.151 & 0.202 & 0.870 \\ \cline{2-6} 
 & hate\_v1 & 0.175 & 0.170 & 0.172 & 0.823 \\ \cline{2-6} 
 & multi\_hate\_v1 & 0.241 & 0.238 & 0.240 & 0.835 \\ \cline{2-6} 
 & hate\_v3 & 0.199 & 0.248 & 0.221 & 0.810 \\ \cline{2-6} 
 & hate\_sex\_v3 & 0.201 & 0.197 & 0.199 & 0.827 \\ \hline
\multirow{6}{*}{Sex} 
 & Mistral\_V3 & 0.581 & 0.699 & 0.634 & 0.830 \\ \cline{2-6} 
 & sex\_v1 & 0.614 & 0.720 & 0.663 & 0.845 \\ \cline{2-6} 
 & multi\_hate\_v1 & 0.589 & 0.668 & 0.626 & 0.832 \\ \cline{2-6} 
 & sex\_v3 & 0.705 & 0.668 & 0.653 & 0.853 \\ \cline{2-6} 
 & hate\_sex\_v3 & 0.628 & 0.664 & 0.664 & 0.886 \\ \hline
\multirow{6}{*}{Political Figure} 
 & Mistral\_V3 & 0.974 & 0.946 & 0.960 & 0.960 \\ \cline{2-6} 
 & hate\_v1 & 0.947 & 0.921 & 0.934 & 0.935 \\ \cline{2-6} 
 & multi\_hate\_v1 & 0.938 & 0.962 & 0.971 & 0.971 \\ \cline{2-6} 
 & hate\_v3 & 0.939 & 0.938 & 0.938 & 0.939 \\ \cline{2-6} 
 & hate\_sex\_v3 & 0.908 & 0.937 & 0.922 & 0.921 \\ \hline
\end{tabular}
\end{table*}
From Table \ref{tab:ablation_results}, we observe the following trends:

Effectiveness of Mixed Data: The use of mixed data (hate\_multi) generally improves the recall and F1 scores across most datasets, indicating that combining different categories of harmful content can enhance the model's ability to generalize and detect diverse harmful content.

Precision Trade-offs: While the recall improves, the precision for some categories decreases slightly. This trade-off suggests that the model becomes more sensitive to identifying harmful content but may also generate more false positives.

Comparison with Single-category Data: Models fine-tuned on single-category data (e.g., hate\_v1, sex\_v1) show high precision but lower recall compared to those fine-tuned on mixed data. This indicates that single-category data supervised fine-tuning might lead to more conservative models that are less likely to generalize to other types of harmful content.

Prompt Versions: The experiments using different prompt versions (v1, v2, v3) show similar performance metrics, indicating that the choice of prompt version has a minimal impact on the overall performance of the fine-tuned models.

The ablation study demonstrates that supervised fine-tuning models with mixed datasets (hate\_multi) can significantly improve recall and F1 scores, enhancing the models' ability to detect various types of harmful content. However, there is a trade-off with precision, requiring careful consideration in applications where false positives are costly. Additionally, the choice of prompt version appears to have a negligible effect on finetuned models performance, suggesting that the primary focus should be on the diversity and quality of the training data.

\end{document}